\documentclass[format=acmsmall, authorversion, review=false, screen=true, nonacm]{acmart}

\usepackage[english]{babel}
\usepackage[T1]{fontenc}
\usepackage[utf8]{inputenc}  
\usepackage[]{hyperref}
\hypersetup{
    colorlinks = true,
    allcolors = blue
}

\usepackage{latexml}

\usepackage{graphicx}
\usepackage{amsmath}
\usepackage{listings}
\usepackage{url}
\usepackage{xcolor}
\usepackage{xspace}
\newcommand{\sroiq}{\ensuremath{\mathcal{SROIQ}}\xspace}

\usepackage{xstring}
\newcommand{\mt}[1]{\texttt{%
\StrSubstitute{#1}{ }{\,}}} %

\newcommand*\justify{%
  \fontdimen2\font=0.4em%
  \fontdimen3\font=0.2em%
  \fontdimen4\font=0.1em%
  \fontdimen7\font=0.1em%
  \hyphenchar\font=`\-%
}

\newcommand{\ttl}[1]{\texttt{\justify #1}}

\pdfoutput=1 %

\begin{document}

\title{gUFO: A Gentle Foundational Ontology for Semantic Web Knowledge Graphs} 

\author{João Paulo A. Almeida}
\email{jpalmeida@ieee.org}
\orcid{0000-0002-9819-3781}
\affiliation{%
  \institution{Federal University of Espírito Santo}
  \department{Conceptual Modeling Research Group (NEMO)}
  \city{Vitória}
  \country{Brazil}
}

\author{Giancarlo Guizzardi}
\email{g.guizzardi@utwente.nl}
\orcid{0000-0002-3452-553X}
\affiliation{%
  \institution{University of Twente}
  \department{Semantics, Cybersecurity \& Services (SCS)}
  \city{Enschede}
  \country{The Netherlands}
}

\author{Tiago Prince Sales}
\email{t.princesales@utwente.nl}
\orcid{0000-0002-5385-5761}
\affiliation{%
  \institution{University of Twente}
  \department{Semantics, Cybersecurity \& Services (SCS)}
  \city{Enschede}
  \country{The Netherlands}
}

\author{Claudenir M. Fonseca}
\email{c.moraisfonseca@utwente.nl}
\orcid{0000-0003-2528-3118}
\affiliation{%
  \institution{University of Twente}
  \department{Semantics, Cybersecurity \& Services (SCS)}
  \city{Enschede}
  \country{The Netherlands}
}

\renewcommand{\shortauthors}{J. P. A. Almeida, G. Guizzardi, T. P. Sales, C. M. Fonseca} %


\begin{abstract}
gUFO is a lightweight implementation of the Unified Foundational Ontology (UFO) suitable for Semantic Web OWL 2 DL applications. UFO is a mature foundational ontology with a rich axiomatization and that has been employed in a significant number of projects in research and industry. Moreover, it is currently in the process of standardization by the International Organization for Standardization as the ISO/IEC CD 21838-5. %
gUFO stands out from other foundational ontology implementations (such as those provided for BFO and DOLCE) given its unique support for a typology of types (operationalizing OntoClean guidelines), its reification patterns for intrinsic and relational aspects, and its support for situations and high-order types. gUFO provides well-founded patterns to address recurrent problems in Semantic Web knowledge graphs. In this paper, we present gUFO with its constituting categories, relations and constraints, discuss how it differs from the original UFO reference ontology, elaborate on its community adoption, and systematically position it in relation to existing OWL-based implementations of popular alternative foundational ontologies.
\end{abstract}


\definecolor{LightGray}{rgb}{0.9,0.9,0.9}
\definecolor{Gray}{rgb}{0.7,0.7,0.7}

\providecommand{\iflatexml}{\iffalse}

\iflatexml

    \lstdefinelanguage{SPARQL}{
    basicstyle=\ttfamily,
    backgroundcolor=\color{LightGray},
    frame=single,
    columns=flexible,
    keepspaces=true,
    breaklines=false,
    sensitive=true,
    tabsize = 2,
    showstringspaces=false,
    morecomment=[s][\color{black}]{<http}{>}, %
    morecomment=[l][\color{gray}]{\#},       %
    morestring=[b][\color{black}]{\"},  %
    keywordsprefix=?,
    classoffset=0,
    keywordstyle=\color{black},
    morekeywords={},
    classoffset=1,
    keywordstyle=\color{black},
    morekeywords={rdf,rdfs,owl,xsd,purl},
    classoffset=2,
    keywordstyle=\color{black}\bfseries,
    morekeywords={
    SELECT,DISTINCT,CONSTRUCT,DESCRIBE,ASK,WHERE,FROM,NAMED,PREFIX,BASE,OPTIONAL,
    FILTER,GRAPH,LIMIT,OFFSET,SERVICE,UNION,EXISTS,NOT,BINDINGS,MINUS, GROUP, BY,
    distinct,construct,describe,ask,where,from,named,prefix,optional,
    filter,graph,limit,offset,service,union,exists,bindings,minus, group, by,
    }
    }
\else
    \lstdefinelanguage{SPARQL}{
    basicstyle=\small\ttfamily,
    backgroundcolor=\color{LightGray},
    rulecolor=\color{LightGray},
    frame=single,
    columns=fullflexible,
    keepspaces=true,
    breaklines=false,
    sensitive=true,
    tabsize = 2,
    showstringspaces=false,
    morecomment=[s][\color{black}]{<http}{>}, %
    morecomment=[l][\color{gray}]{\#},       %
    morestring=[b][\color{black}]{\"},  %
    keywordsprefix=?,
    classoffset=0,
    keywordstyle=\color{black},
    morekeywords={},
    classoffset=1,
    keywordstyle=\color{black},
    morekeywords={rdf,rdfs,owl,xsd,purl},
    classoffset=2,
    keywordstyle=\color{black}\bfseries,
    morekeywords={
    SELECT,DISTINCT,CONSTRUCT,DESCRIBE,ASK,WHERE,FROM,NAMED,PREFIX,BASE,OPTIONAL,
    FILTER,GRAPH,LIMIT,OFFSET,SERVICE,UNION,EXISTS,NOT,BINDINGS,MINUS, GROUP, BY,
    distinct,construct,describe,ask,where,from,named,prefix,base,optional,
    filter,graph,limit,offset,service,union,exists,bindings,minus, group, by,
    }
    }
\fi

\maketitle

\section{Introduction}
\label{sec:intro} 

Foundational ontologies (also termed top-level or upper-level ontologies) are domain-independent and well-founded systems of real-world categories~\cite{schneider10.1007/978-3-540-39451-8_10}. 
They address the most pervasive ontological distinctions underlying language and cognition, with categories to account for objects, events, relationships, properties, situations, etc. 

Over the years, several foundational ontologies have been developed~\cite{arp2015building,Guizzardi2005,Heller2004, masolo2003} 
and applied as a basis for the development of domain ontologies~\cite{Heller2004} and in the revision and redesign of conceptual modeling languages~\cite{Guizzardi2005}. They have been found to leverage foundational patterns to domain ontologies~\cite{gangemi10.1007/11574620_21}, improving their quality~\cite{keet10.1007/978-3-642-21034-1_22}, beyond mere logical consistency. They have also been found useful to improve taxonomies~\cite{Gangemi_Guarino_Masolo_Oltramari_2003} and provide real-world semantics and modeling guidelines for structural conceptual languages~\cite{Guizzardi2005}.

One such ontology is the Unified Foundational Ontology (UFO)~\cite{Guizzardi2005}, which was developed by consistently putting together a number of theories originating from areas formal ontology in philosophy, cognitive science, linguistics and philosophical logics. It currently comprises a number of micro-theories~\cite{ufo_unified_foundational_ontology_2021} addressing objects, their types, their properties, the events they participate in, etc. UFO is a mature foundational ontology with a rich axiomatization~\cite{ufo_unified_foundational_ontology_2021,Guizzardi2005} and that has been employed in a significant number of projects in research and industry \cite{ufo_unified_foundational_ontology_2021,towards_ontological_foundations_for_conceptual_modeling__the_unified_foundational_ontology__ufo__story_2015} (in particular, see Section~5 of \cite{ufo_unified_foundational_ontology_2021} which reports the adoption of UFO in a wide range of domains and applications). %
It is currently approved for registration as an ISO/IEC Draft International Standard (ISO/IEC DIS 21838-5) \cite{iso218385}.

One way of leveraging the benefits of foundational ontologies to domain ontologies is to encode the foundational ontology and a domain ontology that reuses the foundational ontology in the same encoding language. In order to realize this approach for Semantic Web domain ontologies specifically, a number of foundational ontologies have been given Semantic Web (OWL) implementations, such as DOLCE's numerous variant implementations (DOLCE Lite-Plus, DOLCE Ultra Lite-Plus, DOLCE+DnS Ultralite,\footnote{\url{http://www.ontologydesignpatterns.org/ont/dul/DUL.owl}} DOLCE Zero), BFO's implementation in OWL,\footnote{\url{https://github.com/BFO-ontology/BFO-2020}} the OWL version of GFO\footnote{\url{https://www.onto-med.de/ontologies/gfo/}} and PROTON.\footnote{\url{https://ontotext.com/documents/proton/Proton-Ver3.0B.pdf}} These ontologies have aimed at coping with the restricted expressiveness of OWL, and in many cases (such as that of DOLCE, BFO and GFO) are adaptations of more complete versions specified in more expressive formalisms. Often, their authors have ascribed them the qualification `lightweight' in order to reflect the simplifications that were made.

In this paper, we present a lightweight implementation of the Unified Foundational Ontology (UFO) dubbed gUFO. gUFO stands out from other foundational ontology implementations given its unique support for a typology of types (which operationalizes OntoClean guidelines~\cite{Guarino2004ontoclean}); its reification patterns for intrinsic and relational `aspects'; and its support for situations and `high-order' types. The objective of gUFO is to provide a lightweight implementation of UFO suitable for Semantic Web OWL 2 DL~\cite{owl2directsemantics} applications providing well-founded patterns to address recurrent problems in Semantic Web knowledge graphs representation, including qualitative change over time, dynamic/contingent  classification, which are not addressed by the aforementioned foundational ontology implementations.

There are two implications of gUFO being a lightweight implementation of UFO we would like to emphasize. First, gUFO employs the expressive means of OWL 2 DL, instead of the more expressive first-order (modal) logics employed in the original formalization of UFO~\cite{ufo_unified_foundational_ontology_2021}. %
Second, while UFO is an implementation-independent reference ontology, gUFO was explicitly designed with the purpose of providing an implementation artifact to structure an RDF/OWL-based knowledge graph, and, hence, it provides patterns and solutions that shape the resulting knowledge graphs in this particular technological space. 

The `g' in gUFO stands for \textit{gentle} (and at the same time, `gufo' is the Italian word for `owl'.)  The term reveals that ease of use has been a key concern in its design. Because of this, gUFO strives to avoid philosophical jargon and follows closely the community culture and best practices of the Semantic Web, with open documentation,\footnote{\url{https://purl.org/nemo/doc/gufo}} stable  repository,\footnote{\url{https://github.com/nemo-ufes/gufo}} CC BY 4.0 licensing, and adherence to FAIR principles~\cite{jacobsen2020fair} for ontologies~\cite{Garijo2020}.

For background information on the reference ontology on which gUFO is based, including full axiomatization and justification, see  \cite{ufo_unified_foundational_ontology_2021,endurant_types_in_ontology_driven_conceptual_modeling__towards_ontouml_2_0_2018,towards_ontological_foundations_for_conceptual_modeling__the_unified_foundational_ontology__ufo__story_2015,towards_ontological_foundations_for_the_conceptual_modeling_of_events_2013,Guizzardi2005}.

This paper is further structured as follows. Section~\ref{sec:requirements} discuss gUFO's intended uses and requirements. Section~\ref{sec:overview} provides an overview of  gUFO, explaining how it is used as a resource for domain ontology implementers. Some preliminary considerations on notation are also provided. Sections~\ref{sec:taxonomy_of_invididuals} and \ref{sec:taxonomy_of_types} present the gUFO taxonomies applicable to individuals and types respectively, conforming with the online gUFO documentation~\cite{almeida2019gufo}. The various ontology patterns that are leveraged to gUFO-based ontologies are presented throughout these sections. Section~\ref{sec:ufogufo} discusses the relation between the original UFO formalization and the gUFO implementation,  revealing a number of key design decisions and rationale. Section~\ref{sec:constraints} shows a number of SHACL constraints that can be used to verify gUFO-based domain ontologies for their quality beyond basic (OWL-based) logical consistency checks. %
Section~\ref{sec:assessment} reports on the assessment of gUFO from the perspective of consistency, freedom from pitfalls and anti-patterns, as well as FAIRness indicators.
Section~\ref{sec:community-adoption} presents a diverse set of third-party works in the community adopting gUFO in both academia and industry.
Section~\ref{sec:related-work} compares gUFO with other Semantic Web foundational ontology implementations.
Finally, Section~\ref{sec:conclusions} provides concluding remarks and outlines some future work.

\section{Intended Uses and Requirements}
\label{sec:requirements}

gUFO was designed to allow practitioners developing domain ontologies and knowledge graphs in RDF/OWL to profit from foundational distinctions and conceptual patterns defined originally in UFO. 
As such, users are expected to rely on well-established tools and Semantic Web technologies. This includes the ability to import gUFO into tools like Prot\'{e}g\'{e}\footnote{\url{https://protege.stanford.edu}} and TopBraid EDG Studio,\footnote{\url{https://www.topquadrant.com/doc/latest/topbraid_edg_studio}} where they can directly refer to imported classes and properties.
Moreover, users are expected to employ OWL~2~DL~\cite{owl2directsemantics} reasoners and SHACL \cite{shacl} tools at design- and/or runtime to enable consistency checks, inferencing, and quality assessment of domain ontology and conforming knowledge graphs.
As such, the technological space used by intended gUFO users (supporting RDF/OWL knowledge graphs) is significantly different from that of the automated theorem provers and model finders used in the development of the original UFO formalization~\cite{ufo_unified_foundational_ontology_2021,towards_ontological_foundations_for_the_conceptual_modeling_of_events_2013}, with consequences for gUFO (as discussed in Section \ref{sec:ufogufo}). 

Among the intended users, we also include conceptual modelers who rely on the OntoUML language \cite{endurant_types_in_ontology_driven_conceptual_modeling__towards_ontouml_2_0_2018, Guizzardi2005}, a UFO-based lightweight extension of the UML class diagram language.
These users can leverage gUFO to operationalize and validate their ontologies, constructed as conceptual models in a diagrammatic syntax.
The shared UFO foundation behind gUFO and OntoUML enables a straightforward translation from diagrams to knowledge graphs, which is facilitated by an automatic transformation available in OntoUML tooling~\cite{ontology_driven_conceptual_modeling_as_a_service_2021}.

In order to facilitate potential reuse of gUFO by designers of domain ontologies, a number of decisions have been taken, which include: (i) the avoidance of philosophical jargon (pervasive in other foundational ontologies), while still leveraging well-founded distinctions in the philosophical literature;
(ii) the availability of extensive documentation, including Aristotelian-style definitions and examples embedded in OWL annotations;
(iii) the adoption of FAIR best practices \cite{Garijo2020} (stable permanent IRI, use of metadata to indicate permissible licensing, versioning, registration of prefix);
(iv) the offering of a single variant, differently from other foundational ontologies which have been given multiple implementations (such as DOLCE).

Similarly to UFO, gUFO is intended to provide conceptual support for:

\begin{itemize}
\item the distinctions between individuals and types; between objects, events and situations; 
\item the characterization of entities through their various aspects (both intrinsic and relational);
\item the reification of relational aspects, which gives a focal point for relationships as key notions for domain abstraction;
\item the treatment of events and the participation of objects in events, including their creation and destruction; 
\item treatment of part-whole relations, for both objects and events.
\end{itemize}

As we show throughout the paper, gUFO can be used in varying degrees of sophistication depending on the use case. This ranges from a minimalist use of gUFO classes and properties directly in knowledge graphs (without designing a domain-specific ontology), to the design of well-founded domain ontologies, which can benefit from (i) gUFO-based design patterns, (ii) a rich taxonomy of types and (iii) support for higher-order classes. 

The ontology design patterns aim to address a number of problems in the representation of knowledge graphs in RDF/OWL. These can be employed in varying degrees of complexity, depending on the level of sophistication required in a domain ontology implementation. Patterns at increasing levels of detail are provided for qualities (such as the mass of a physical object, the height of a person) and also for relationships (marriages, enrollments). Patterns based on situations are also provided to address, when users require, temporal aspects of objects and events, changes in classification, changes in values of properties and in part-whole relations. 

While OWL does not distinguish different types of classes, gUFO introduces explicitly the taxonomy of types of UFO, classifying types through their metaproperties such as rigidity, sortality and external dependence~\cite{Guizzardi2005}. As discussed in~\cite{Guarino2004ontoclean}, these metaproperties can be used to guide the production of higher quality taxonomies. The use of gUFO enables the expression of said metaproperties for OWL classes, and the implementation of (SHACL) constraints which allow for their automated verification, not unlike what is offered for OntoUML users~\cite{ontology_driven_conceptual_modeling_as_a_service_2021,endurant_types_in_ontology_driven_conceptual_modeling__towards_ontouml_2_0_2018}. (See Sections~\ref{sec:ufogufo} and \ref{sec:constraints} for motivation and implementation of these constraints.)

gUFO users can also benefit from the definition of domain high-order classes~\cite{foxvog2005instances} (also known as \textit{powertypes} in the conceptual modeling literature \cite{odell1994power}). These are present in domains in which we are concerned not only with types of individuals, but also with types of types. Examples include biological taxonomy (where animal species and taxa in general are in the domain of interest), product types (with product models, categories), law (with penal types), manufacturing (with types of materials in bills). Some of us have shown in \cite{evidence_of_large_scale_conceptual_disarray_in_multi_level_taxonomies_in_wikidata_2023} that these \textit{multi-level} representation phenomena are pervasive in knowledge graphs such as Wikidata, and that they are, at the same time, most challenging without support from specialized rules~\cite{multi_level_conceptual_modeling_theory_language_and_application_2021}. The use of gUFO enables the automatic verification of some of these rules.

\section{Overview and Preliminaries}
\label{sec:overview}

gUFO includes 51 classes, 40 object properties and 7 data properties. It is specified in OWL~2~DL. %
Its elements are reused in gUFO-based (domain) ontologies, which inherit from gUFO its domain-independent distinctions. Throughout this paper we use the prefix \mt{gufo:} for gUFO elements\footnote{Denoting \url{http://purl.org/nemo/gufo\#}.} (e.g., \mt{gufo:Object}, \mt{gufo:Event}) and the default prefix \mt{:} for the elements of an example ontology reusing gUFO (e.g., \mt{:Car}, \mt{:RockConcert}). The prefixes \mt{rdfs:} and \mt{xsd:} are used as usual.\footnote{Denoting \url{http://www.w3.org/2000/01/rdf-schema\#} and \url{http://www.w3.org/2001/XMLSchema\#}.} We use Turtle as a syntax for RDF serialization for illustrative purposes.

A key feature of gUFO is that it includes two disjoint taxonomies: one with classes whose instances are \textit{individuals}, i.e., entities that do not have instances (such as Paul McCartney and the Earth), and another with classes whose instances are \textit{types} (such as `Person' and `Planet'). Classes in the taxonomy of individuals include \mt{gufo:Object},  \mt{gufo:Quality}, \mt{gufo:Situation}, \mt{gufo:Event} (to name a few), whose purpose is to establish the ontological nature of the individuals they classify. Classes in the taxonomy of types include \mt{gufo:Kind}, \mt{gufo:Phase}, \mt{gufo:Category} (again, to name a few), whose purpose is to provide a typology of types that settles the metaproperties (rigidity, sortality, external dependence~\cite{guarino2000_10.1007/3-540-39967-4_8}) of the types they classify.

Figure \ref{fig:gufo-taxonomies} shows an overview of gUFO, using a UML class diagram for visualization purposes. (Disjoint and complete generalization sets in the diagram correspond to the superclass being declared as disjoint union of the subclasses in that set; disjoint generalization sets correspond to pairwise disjointness declarations for the subclasses in the set). The taxonomy of individuals is shown in the right-hand side, and the taxonomy of types is shown on the left-hand side. A few classes have been omitted for brevity \cite{almeida2019gufo}.

\iflatexml
    \begin{figure}[ht]
    \centering
    \includegraphics[width=\textwidth]{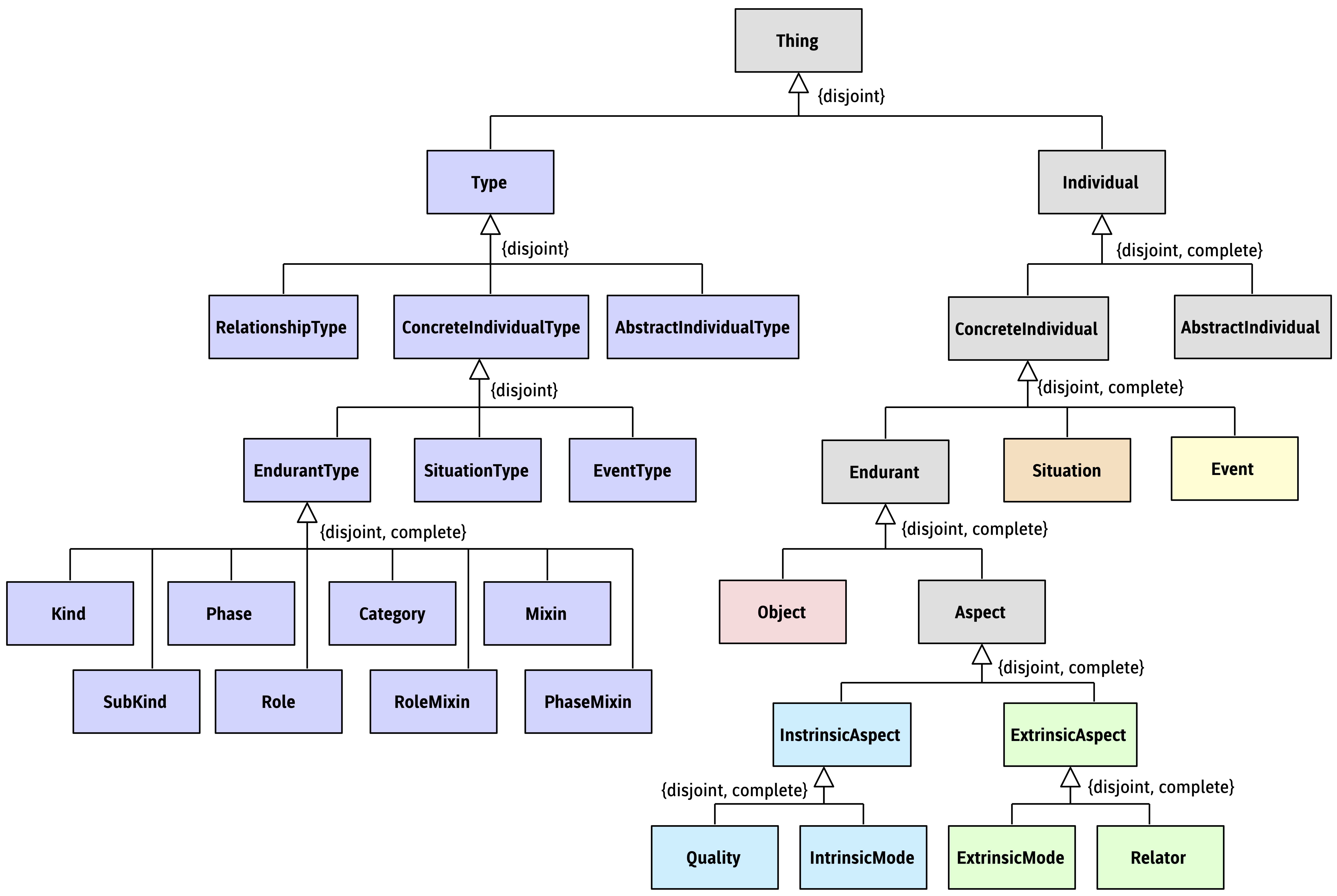}
    \caption{gUFO taxonomy overview.}
        \Description{gUFO taxonomy overview.}
    \label{fig:gufo-taxonomies}
    \end{figure}
\else
    \begin{figure}[ht]
    \centering
    \includegraphics[width=0.9\textwidth]{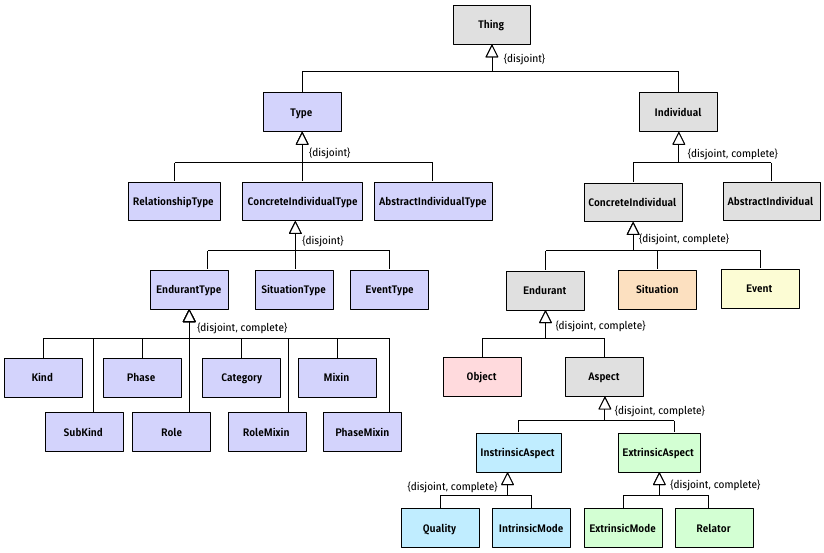}
    \caption{gUFO taxonomy overview.}
        \Description{gUFO taxonomy overview.}
    \label{fig:gufo-taxonomies}
    \end{figure}
\fi

Reuse of gUFO consists in instantiating and/or specializing its classes, object properties and data properties, inheriting from them the domain-independent distinctions of UFO. Considering the two taxonomies, the following usage scenarios are applicable:

\begin{enumerate}
    \item A reusing knowledge graph declares individuals \textit{instantiating gUFO classes in the taxonomy of individuals}, e.g., \ttl{:Earth a gufo:Object} and \ttl{:WorldCup1970Final a gufo:Event}.
    \item A reusing ontology declares classes \textit{specializing gUFO classes in the taxonomy of individuals}, e.g., \ttl{:Planet rdfs:subClassOf gufo:Object} and \ttl{:SoccerMatch rdfs:subClassOf gufo:Event}.
    \item A reusing ontology declares classes with specific metaproperties \textit{instantiating gUFO classes in the taxonomy of types}. For example, \ttl{:Planet a gufo:Kind\footnote{These type distinctions, e.g., between kinds, phases, etc., are discussed in Section \ref{sec:taxonomy_of_types}.}} and \ttl{:Child a gufo:Phase}.
    \item A reusing ontology declares high-order classes \textit{specializing gUFO classes in the taxonomy of types}. For example, \ttl{:PersonPhase rdfs:subClassOf gufo:Phase}.
\end{enumerate}

Designers of gUFO-based ontologies may combine these scenarios. For example, scenarios 2 and 3 are often employed in combination as shown in Listing \ref{lst:scenarios}, which defines a \mt{Person} class that specializes \mt{gufo:Object} and instantiates \mt{gufo:Kind}. It settles the ontological nature of instances of \mt{Person} and it defines that, as an instance of \mt{gufo:Kind}, \mt{Person} is both a rigid and sortal type, hence applying necessarily to its instances, and providing them a uniform principle of identity. 

\begin{lstlisting}[language=SPARQL, label=lst:scenarios, caption=A fragment of an ontology specializing a class in the taxonomy of individuals and instantiating a class in the taxonomy of types.]
:Person a owl:Class ;
        rdfs:subClassOf gufo:Object ;
        a gufo:Kind .
\end{lstlisting}

\section{Taxonomy of Individuals}
\label{sec:taxonomy_of_invididuals}

The topmost distinction in the taxonomy of individuals partitions them into \textit{concrete} and \textit{abstract individuals}. A \mt{gufo:ConcreteIndividual}, differently from a \mt{gufo:AbstractIndividual}, is one that exists in space-time. Concrete individuals comprise \textit{objects} (Mount Everest, John, his car, the Brazilian 1988 Constitution), reified \textit{aspects} of concrete individuals (John's height, his service agreement with Amazon, Inc.), \textit{events} (the 1986 Mexico City Earthquake, the 2009 United Nations Climate Change Conference) and \textit{situations} (the situation in which John weighs 80 kilograms, the situation in which Mary is a professor).

Objects and aspects are termed \textit{endurants}, which are concrete individuals that endure in time and may change qualitatively while keeping their identity. A \mt{gufo:Aspect} is a characteristic or trait of a concrete individual that is itself conceived as an individual. Treating aspects as endurants (i.e., reifying aspects) allows us to consider the properties of aspects themselves, and account for their change in time.

\subsubsection*{Properties for Concrete Individuals}

gUFO includes a number of data and object properties that can be used with concrete individuals. For example, temporal aspects of concrete individuals can be captured with the data properties \mt{gufo:hasBeginPointInXSDDate}, \mt{gufo:hasBeginPointInXSDDate\-TimeStamp},  \mt{gufo:hasEndPointInXSDDate} and \mt{gufo:hasEndPointInXSDDateTimeStamp}, depending on the chosen level of granularity. In the case of objects (and aspects), these properties determine the time point when the object (or aspect) comes into existence or ceases to exist. In the case of events, these properties determine the time point when the event starts to take place or when it ends. In the case of situations, the properties determine the time point when the situation begins and ceases to hold. Temporal instants may also be reified using OWL-Time~\cite{owltime} \mt{time:Instant} in which case the \mt{gufo:hasBeginPoint} and \mt{gufo:hasEndPoint} object properties are applicable. See OWL-Time~\cite{owltime} for details concerning \mt{time:Instant}, including support for Allen relations, temporal reference systems, temporal precision.

Listing \ref{lst:earthquake} shows the use of these properties to represent that the 1985 Mexico City Earthquake started at 13:17:50 (UTC) on the 19th Sept. 1985 (instantiating \mt{gufo:Event} in line with usage scenario 1).

\begin{lstlisting}[language=SPARQL,label=lst:earthquake,caption=A fragment of a knowledge graph declaring an event with a temporal property.]
:1985MexicoCityEarthquake a gufo:Event;
        gufo:hasBeginPointInXSDDateTimeStamp "1985-09-19T13:17:50Z"^^xsd:dateTimeStamp .
\end{lstlisting}

Listing \ref{lst:earthquake2} shows Earthquake as be a sub-class of \mt{gufo:Event}, and hence, the applicable object and data properties can be used with particular instances of earthquake (subclassing \mt{gufo:Event} in line with usage scenario 2).

\begin{lstlisting}[language=SPARQL,label=lst:earthquake2,caption=A fragment of an ontology declaring an event type and an instance of it with a temporal property.]
:Earthquake a owl:Class ;
            rdfs:subClassOf gufo:Event .

:1985MexicoCityEarthquake a :Earthquake ;
        gufo:hasBeginPointInXSDDateTimeStamp "1985-09-19T13:17:50Z"^^xsd:dateTimeStamp .
\end{lstlisting}                        
                        
\clearpage
\subsection {Objects}

Objects are further classified in gUFO according to the way they are structured into parts, leading to the following subclasses: \mt{gufo:FunctionalComplex}, \mt{gufo:Collection} and \mt{gufo:Quantity}.

A \mt{gufo:FunctionalComplex} is a complex \mt{gufo:Object} whose parts (components) play different roles in its composition, including most ordinary objects. For example, a person could be considered a \mt{gufo:FunctionalComplex} with their various organs (heart, brain, lungs, etc.) playing different roles in the context of their body.

A \mt{gufo:Collection} is a complex gufo:Object whose parts (the members of the collection) have a uniform structure (i.e., members are conceived as playing the same role in the collection). Examples include a deck of cards, a pile of bricks, a forest (conceived as a collection of trees), a group of people. Collections may have a variable or fixed membership (subclasses \mt{gufo:VariableCollection} and \mt{gufo:FixedCollection}).

A \mt{gufo:Quantity} is a complex \mt{gufo:Object} that is a maximally-connected portion of stuff. A \mt{gufo:Quantity} has a fixed constitution, and thus, removing or adding a sub-quantity would result in a different quantity. Examples include a particular portion of wine in a wine tank (as opposed to any generic 1L of wine), a particular lump of clay, the particular gold that constitutes a wedding ring.

The relations between a part and its whole are captured with the \mt{gufo:isObjectProperPartOf} property and its sub-properties, depending on the types of parts and complex objects involved:
\begin{itemize}
    \item \mt{gufo:isComponentOf}, when the part is a component of a functional complex;
    \item \mt{gufo:isCollectionMemberOf}, when the part is a member of a collection;
    \item \mt{gufo:isSubCollectionOf}, when the part is a collection containing a proper subset of all the members of another collection; and,
    \item \mt{gufo:isSubQuantityOf},  when the part and the whole are quantities.
\end{itemize}

Listing \ref{lst:johnsbrain} illustrates the usage of \mt{gufo:isComponentOf} with John's brain as a component of his.

\begin{lstlisting}[language=SPARQL,label=lst:johnsbrain,caption=A fragment of an ontology declaring John's brain as a component of his.]
:John a :Person .

:Brain a owl:Class ; 
       rdfs:subClassOf gufo:Object .

:JohnsBrain a :Brain ;
            gufo:isComponentOf :John .
\end{lstlisting}

\subsection {Aspects}

Aspects in gUFO are divided into intrinsic aspects (qualities and intrinsic modes) and extrinsic aspects (relators and extrinsic modes). A \mt{gufo:IntrinsicAspect} depends on a single concrete individual in which it inheres. Examples include intrinsic physical aspects, such as the Moon's mass, Lassie's fur color; the fragility of John Lennon's glasses; mental dispositions, such as Bob's math skills, his belief that the number one is odd. Intrinsic aspects are divided into: (i) qualities, when the aspect is projected into one or more suitable value spaces (for example, Bob's weight or the height of the Statue of Liberty), and (ii) intrinsic modes, which are not given a direct value (for example, Bob's belief that the Eiffel Tower is in Paris or his capability to speak Japanese).

\subsubsection{Qualities}

The most straightforward way to describe intrinsic aspects of entities in an RDF/OWL setting is to employ data properties. For example, the fragment in Listing \ref{lst:massdataproperty} defines a data property to represent the mass of a physical object in kilograms and declare a value for the Moon's mass.

\begin{lstlisting}[language=SPARQL,label=lst:massdataproperty,caption=A fragment of an ontology showing the mass of the Moon with a plain data property.]
:hasMassInKilograms a owl:DatatypeProperty ;
                    rdfs:domain :PhysicalObject;
                    rdfs:range xsd:double .

:Moon a :PhysicalObject ;
      :hasMassInKilograms "7.34767309E22"^^xsd:double .
\end{lstlisting}

While this is an adequate solution in many settings, a more sophisticated representation may be required, which  can be obtained in gUFO by reifying an aspect as a concrete individual (as an instance of \mt{gufo:Aspect}). 

The fragment in Listing \ref{lst:moonreifiedquality} shows an example of definition of a quality type (\mt{:Mass}) by specializing \mt{gufo:Quality}. As an intrinsic aspect, a \mt{gufo:Quality} \mt{gufo:inheresIn} a \mt{gufo:Concrete\-Individual}. For example, the Moon's mass inheres in the Moon. Further, a quality may be given a value with the \mt{gufo:hasQualityValue} data property. 

\begin{lstlisting}[language=SPARQL,label=lst:moonreifiedquality,caption=A fragment of an ontology showing the reification of a quality.]
:Mass a owl:Class ;
      rdfs:subClassOf gufo:Quality .

:MoonsMass a :Mass ;
           gufo:inheresIn :Moon ;
           gufo:hasQualityValue "7.34767309E22"^^xsd:double .
\end{lstlisting}

A user may define specialized sub-properties of \mt{gufo:inheresIn} in order to determine specific domain and range applicable to qualities of a certain type (cf. Listing \ref{lst:massOf}).

\begin{lstlisting}[language=SPARQL,label=lst:massOf,caption=A fragment of an ontology showing an object property to determine the mass of a physical object.]
:massOf a owl:ObjectProperty ;
        rdfs:subPropertyOf gufo:inheresIn ;
        rdfs:domain :Mass ;
        rdfs:range :PhysicalObject .
\end{lstlisting}
        
A user may also define sub-properties of \mt{gufo:hasQualityValue}, for example with different ways to quantify a quality. Treating a quality (such as the Moon's mass) as a reified entity, we can give it a value in different units (e.g., in kilograms and in short tons), see Listing~\ref{lst:units}. This structure differs from the plain usage of various data properties as the reification makes it clear we are talking about the very same mass. %

\begin{lstlisting}[language=SPARQL,label=lst:units,caption=A fragment of an ontology showing a quality with different measurements units.]
:massInKilograms a owl:DatatypeProperty ;
                 rdfs:subPropertyOf gufo:hasQualityValue ;
                 rdfs:domain :Mass ;
                 rdfs:range xsd:double .

:massInShortTons a owl:DatatypeProperty ;
                 rdfs:subPropertyOf gufo:hasQualityValue ;
                 rdfs:domain :Mass ;
                 rdfs:range xsd:double .

:MoonsMass a :Mass ;
           :massOf :Moon ;
           :massInKilograms "7.34767309E22"^^xsd:double ;
           :massInShortTons "8.099423160000001E19"^^xsd:double .
\end{lstlisting}

More advanced reification patterns are also defined in gUFO, involving the reification of quality values themselves. An object property called \mt{gufo:hasReifiedQualityValue} is provided for those cases in which an instance of \mt{gufo:QualityValue} is used instead of a literal to provide the value of a quality. Reifying a \mt{gufo:QualityValue} may be useful in the case of enumerated values (for example, the various sizes of t-shirts), nominal qualities (for example, ethnicity or gender) or for quality values that are defined in terms of a multidimensional quality structure (such as color conceived in terms of red, green and blue components or hue, saturation and brightness).

\subsubsection{Intrinsic Modes}

Intrinsic modes, unlike qualities, are not conceived of in terms of their projection into value spaces. The fragment in Listing \ref{lst:intrinsicmode1} shows an example of a sub-class of \ttl{gufo:IntrinsicMode} and the corresponding sub-property of \ttl{gufo:inheresIn} establishing that a headache inheres in a person.
\begin{lstlisting}[language=SPARQL,label=lst:intrinsicmode1,caption=A fragment of an ontology showing an example of intrinsic mode class.]
:Headache a owl:Class ;
          rdfs:subClassOf gufo:IntrinsicMode .

:headacheOf a owl:ObjectProperty ;
            rdfs:subPropertyOf gufo:inheresIn ;
            rdfs:domain :Headache ;
            rdfs:range :Person .
\end{lstlisting}
            
Since a \ttl{gufo:IntrinsicMode} is a \ttl{gufo:ConcreteIndividual}, we may track its temporal properties, as well as ascribe qualities (or other aspects) to it (such as intensity, location where it is felt). Listing \ref{lst:intrinsicmode2} considers the \textit{intensity} of a headache as a quality in itself with an associated Verbal Rating Scale (with 1 corresponding to ``Mild pain'', 2 corresponding to ``Moderate pain'', 3 corresponding to ``Severe''):

\begin{lstlisting}[language=SPARQL,label=lst:intrinsicmode2,caption=A fragment determining of a quality and its value for an intrinsic mode.]
:johnsHeadache a :Headache ;
               :headacheOf :John ;
               gufo:hasBeginPointInXSDDateTimeStamp 
                            "2019-11-19T14:14:50Z"^^xsd:dateTimeStamp .

:HeadacheIntensity a owl:Class ;
                   rdfs:subClassOf gufo:Quality .

:johnsHeadacheIntensity a :HeadacheIntensity ;
                        gufo:inheresIn :johnsHeadache ;
                        gufo:hasQualityValue "2"^^xsd:nonNegativeInteger .
\end{lstlisting}

\subsubsection{Relators}

While an intrinsic aspect depends on a single concrete individual, an extrinsic (or ``relational'') aspect depends on more than one concrete individual. They are particularly useful in the reification of relationships, through \textit{relators} (instances of \mt{gufo:Relator}), e.g., John and Mary's marriage, Mary's employment contract at NASA. 

A relator \mt{gufo:mediates} concrete individuals. The fragment in Listing \ref{lst:mediates} shows an example of a sub-class of \mt{gufo:Relator} (\mt{:Marriage}) and the corresponding sub-property of \mt{gufo:mediates} establishing that persons are mediated through marriage. 

\begin{lstlisting}[language=SPARQL,label=lst:mediates,caption=A fragment of an ontology with the reification of marriages.]
:Marriage a owl:Class ;
          rdfs:subClassOf gufo:Relator .

:marriageInvolves a owl:ObjectProperty ;
                  rdfs:subPropertyOf gufo:mediates ;
                  rdfs:domain :Marriage ;
                  rdfs:range :Person .
\end{lstlisting}                
                
The fragment in Listing \ref{lst:johnmarrymarriage} instantiates \mt{:Marriage} and establishes the related persons (John and Mary).

\begin{lstlisting}[language=SPARQL,label=lst:johnmarrymarriage,caption=The reification of John and Mary's marriage.]
:John a :Person .
:Mary a :Person .

:JohnMarysMarriage a :Marriage ;
                   :marriageInvolves :John , :Mary .
\end{lstlisting}                                  

Since a relator is a concrete individual, we may also track its temporal properties (e.g., begin and end date of employment), which is not obtained directly with a plain object property. We may also ascribe other aspects to it (e.g., salary in the scope of the employment). Again, this is not supported directly with a plain object property. The use of relators is a well-founded means to approach the `Qualified Relation' pattern~\cite{dodds_davis_2022}, and naturally supports both `N-Ary Relations'~\cite{dodds_davis_2022,nary_w3c} as well as Anadyc relations, i.e., relations whose arity can vary for each of their instances (e.g., the `playing with' relation among children can be binary, ternary, quaternary, etc., in different cases)~\cite{Guizzardi2005}.

A fuller application of the relator pattern can also make explicit the roles that are played in the scope of a relationship, not only by creating subproperties of \mt{gufo:mediates} as shown above, but also by creating instances of \mt{gufo:Role} (see \cite{almeida2019gufo} and the taxonomy of types in Section \ref{sec:taxonomy_of_types}).

\subsubsection{Extrinsic Modes}

Extrinsic aspects can also be reified one-sided relationships, e.g., John's admiration for Obama (which depends on Obama but does not characterize him), in which case they are termed \textit{extrinsic modes}. They can also be used to reveal parts of relators, e.g., John's rights towards Amazon, Inc. (in the scope of a service agreement) and Amazon's reciprocal duties, Amazon's rights towards John, John's reciprocal duties. This is illustrated in Listing \ref{lst:amazonexample}, which also reviews the use of the \ttl{gufo:externallyDependsOn} and  \ttl{gufo:isAspectProperPartOf} properties.

\begin{lstlisting}[language=SPARQL,label=lst:amazonexample,caption={A fragment with an example of extrinsic mode, zooming in on the relationship between John and Amazon, Inc.}]
:AmazonInc a :Organization .

:ServiceAgreement a owl:Class ;
                  rdfs:subClassOf gufo:Relator .

:JohnAmazonAgreement a :ServiceAgreement ;
                     :mediates :AmazonInc , :John .

:JohnsRightToServiceProvisioning a gufo:ExtrinsicMode ;
                gufo:inheresIn :John ;
                gufo:externallyDependsOn :AmazonInc ;
                gufo:isAspectProperPartOf :JohnAmazonAgreement .

:AmazonsDutyToProvideService a gufo:ExtrinsicMode ;
                gufo:inheresIn :AmazonInc ;
                gufo:externallyDependsOn :John ;
                gufo:isAspectProperPartOf :JohnAmazonAgreement .
\end{lstlisting}

\subsection{Events}

A \mt{gufo:Event} is a \mt{gufo:ConcreteIndividual} that `occurs' or `happens' in time. They may be instantaneous or durative. Events are those ``things that happen to or are performed by'' endurants~\cite{sep-events}.

Examples include actions and processes, such as a business meeting, a communicative act, a soccer match, a goal kick, the clicking of a mouse button; as well as natural occurrences such as an earthquake, the fall of the meteor that caused the extinction of the dinosaurs.

The fragment in Listing \ref{lst:events} shows examples of sub-classes of \mt{gufo:Event} (\mt{:NaturalDisaster}, \mt{:Earthquake}, \mt{:Tsunami}, \mt{:SoccerMatch}), and an instance of \mt{:SoccerMatch} (\mt{:WorldCup1970Final}).

\begin{lstlisting}[language=SPARQL,label=lst:events,caption=Examples of classes of events and an instance of event.]
:NaturalDisaster a owl:Class ;
                 rdfs:subClassOf gufo:Event .

:Earthquake a owl:Class ;
            rdfs:subClassOf :NaturalDisaster ;
            owl:disjointWith :Tsunami .

:Tsunami a owl:Class ;
         rdfs:subClassOf :NaturalDisaster .

:SoccerMatch a owl:Class ;
             rdfs:subClassOf gufo:Event .

:WorldCup1970Final a :SoccerMatch .
\end{lstlisting}

The implementation includes an object property to declare historical dependence between events~\cite{towards_ontological_foundations_for_the_conceptual_modeling_of_events_2013}. For example, the \mt{:WorldCup1970Final} depends historically on the two semifinals (Listing \ref{lst:eventsdependence}). When used between events, historical dependence encompasses causation (when the event is caused by the other), but also other cases where there is dependence but not causation (when the event brings about a situation that is either insufficient or more than sufficient to trigger the historically dependent event). For example, Real Madrid's goal in the 60th minute of the 2016 FIFA Club World Cup Final is historically dependent on (and in this case caused by) a penalty kick by Cristiano Ronaldo. The penalty kick itself is historically dependent on (but not caused by) a penalty (the occurrence of the penalty is necessary but not sufficient to cause the penalty kick as authorization of the referee is required). Historical dependence is transitive. Hence, in the example above, Real Madrid's goal is historically dependent on the penalty. In summary, historical dependence is stronger than mere temporal ordering but not necessarily as strong as causation.
\clearpage
\begin{lstlisting}[language=SPARQL,label=lst:eventsdependence,caption=Historical dependence between events.]
:WorldCup1970Final  gufo:historicallyDependsOn 
                    :BrazilUruguayWorldCup1970SemiFinal .
:WorldCup1970Final  gufo:historicallyDependsOn 
                    :ItalyWestGermanyWorldCup1970SemiFinal .
\end{lstlisting}

The relations between objects and events may be captured with \mt{gufo:participatedIn}, \mt{gufo:was\-CreatedIn}, \mt{gufo:wasTerminatedIn}. Sub-properties of these object properties may be created to establish a particular domain and range. The  fragment in Listing \ref{lst:eventparticipation} establishes the participation of players in soccer matches, and defines that Pelé participated in the 1970 World Cup Final.

\begin{lstlisting}[language=SPARQL,label=lst:eventparticipation,caption=Participation of objects in events.]
:SoccerMatchPlayer a owl:Class ;
                   rdfs:subClassOf :Person .

:participatedInMatch a owl:ObjectProperty ;
                     rdfs:subPropertyOf gufo:participatedIn ;
                     rdfs:domain :SoccerMatchPlayer ;
                     rdfs:range :SoccerMatch .

:Pele a :Person ;
      :participatedInMatch :WorldCup1970Final .
\end{lstlisting}

Part-whole relations between events can be represented with the \mt{gufo:isEventProperPartOf} object property (e.g., \ttl{:WorldCup1970Final gufo:isEventProperPartOf :WorldCup1970}). Further, the \mt{gufo:Participation} subclass of \mt{gufo:Event}  can be used for the cases in which we want to represent explicitly each participation as a part of an event with multiple participations~\cite{towards_ontological_foundations_for_the_conceptual_modeling_of_events_2013}.

An event can also be related to the endurants that are created or terminated in it. For example, John and Mary's marriage was brought into existence in their wedding ceremony (Listing \ref{lst:wedding}).

\begin{lstlisting}[language=SPARQL,label=lst:wedding,caption=A wedding resulting in a marriage.]
:JohnMarysMarriage a :Marriage ;
                   :marriageInvolves :John , :Mary ;
                   gufo:wasCreatedIn :JohnMarysWedding.

:JohnMarysWedding a gufo:Event ;
                  gufo:hasBeginPointInXSDDate "2001-12-12"^^xsd:date ;
                  gufo:hasEndPointInXSDDate "2001-12-12"^^xsd:date .
\end{lstlisting}

The \mt{gufo:manifestedIn} property can be used to identify specific aspects that manifest themselves in an event. For example, during the Space Shuttle Challenger (OV-099) launch on January 28, 1986, a flaw in the seals of one of its rocket boosters led to catastrophic failure. Such a flaw can be considered an aspect of the seal, which was manifested in that tragic event (Listing \ref{lst:manifestation}). The reification of aspects such as vulnerabilities and other dispositions is an important feature of risk modeling and management~\cite{the_common_ontology_of_value_and_risk_2018}. Likewise, the reification of aspects such as capabilities and competences is an important feature of capability and competence modeling and  management~\cite{modeling_competences_in_enterprise_architecture_from_knowledge_skills_and_attitudes_to_organizational_capabilities_2024,modeling_resources_and_capabilities_in_enterprise_architecture__a_well_founded_ontology_based_proposal_for_archimate_2015}.

\begin{lstlisting}[language=SPARQL,label=lst:manifestation,caption=Manifestation of an aspect in an event.]
:Challenger a gufo:FunctionalComplex ;
            gufo:wasTerminatedIn :Challengers10thLaunch .

:Challengers10thLaunch a gufo:Event 
                       gufo:hasBeginPointInXSDDate "1986-01-28"^^xsd:date ;
                       gufo:hasEndPointInXSDDate "1986-01-28"^^xsd:date .
    
:ChallengerRightBoosterSeal a gufo:FunctionalComplex ;
                            gufo:isComponentOf :Challenger .

:ChallengerRightBoosterSealFlaw a gufo:IntrinsicMode ;
                                gufo:inheresIn :ChallengerRightBoosterSeal ;
                                gufo:manifestedIn :Challengers10thLaunch .
\end{lstlisting}

\subsection{Situations}
\label{sec:situations}

Situations can be used to represent certain configurations of entities that can be comprehended as a whole, and that are contingent state of affairs, i.e., configurations that hold for a particular time interval but that could be otherwise. The various subclasses of \mt{gufo:Situation} are used to represent change in an otherwise ``immutable'' knowledge graph. This includes: (i) the attribution of value to mutable qualities (such as a person's weight), (ii) the temporary instantiation of non-rigid types (e.g., as someone is a child at one point in time and a teenager later), (iii) the temporary participation in part-whole relations for replaceable parts (such as a car's tires), (iv) the temporary participation in constitution relations (e.g., a statue being constituted by different portions of matter at different points in time), and (v) the temporary participation in mutable relations (e.g., the temporary inflation rate difference between two countries). Other subclasses may be created to capture domain-specific notions such as \mt{:HazardousSituation}, \mt{:PersonHasFever}.

For example, a \mt{gufo:QualityValueAttributionSituation} is used in a pattern when it is necessary to indicate the period of time in which the quality value attribution holds, and track changes in quality value. For example, consider tracking John's weight (technically, his mass) over the years (Listing \ref{lst:situation}). We declare various instances of \mt{gufo:QualityValueAttributionSituation} and declare John to stand in these different situations with the \mt{gufo:standsInQualifiedAttribution} object property. This solves a recurrent problem in Semantic Web representation (the representation of change in time).

\begin{lstlisting}[language=SPARQL,label=lst:situation,caption=Situations allowing the representation of the change of a quality value in time.]
:JohnWeighs80Kgin2015 a gufo:QualityValueAttributionSituation ;
                      gufo:concernsQualityType :Mass ;           
                      gufo:concernsQualityValue "80.0"^^xsd:double ;
                      gufo:hasBeginPointInXSDDate "2015-01-01"^^xsd:date ;
                      gufo:hasEndPointInXSDDate "2015-12-31"^^xsd:date . 

:JohnWeighs70Kgin2018 a gufo:QualityValueAttributionSituation ;
                      gufo:concernsQualityType :Mass ;
                      gufo:concernsQualityValue "70.0"^^xsd:double ;
                      gufo:hasBeginPointInXSDDate "2018-01-01"^^xsd:date ;
                      gufo:hasEndPointInXSDDate "2018-12-31"^^xsd:date .

:John gufo:standsInQualifiedAttribution :JohnWeighs80Kgin2015 ;
      gufo:standsInQualifiedAttribution :JohnWeights70Kgin2018 .
\end{lstlisting}

The \mt{gufo:concernsQualityValue} data property is used to indicate a quality value attributed to the \mt{gufo:Endurant} standing in the situation, and the \mt{gufo:concernsQualityType} is used to identify the quality type (sub-class of \mt{gufo:Quality}) whose value is attributed in the \mt{gufo:QualityValueAttributionSituation}. %

\section{Taxonomy of Types}
\label{sec:taxonomy_of_types}

The taxonomy of types can be used to provide additional information about classes in gUFO-based ontologies. The most abstract classes in the taxonomy of individuals mostly reflect the taxonomy of individuals. For example, a \mt{gufo:AbstractIndividualType} is a \mt{gufo:Type} whose instances are abstract individuals (e.g., \mt{:NaturalNumber}, \mt{:Set}, \mt{:Proposition}), a \mt{gufo:EndurantType} is a \mt{gufo:Type} whose instances are objects and aspects (e.g., \mt{:Person}, \mt{:Marriage}, \mt{:Color}), a \mt{gufo:\-EventType} is a \mt{gufo:Type} whose instances are events (e.g., \mt{:Earthquake}, \mt{:MusicalPerformance}), and so on.

\subsection{Endurant Types}

The taxonomy of endurant types is more detailed, in order to qualify the ways in which an endurant type applies to their instances. For example, Listing~\ref{lst:endurantypes} declares \mt{:Person} to be a rigid sortal (a type that applies necessarily to its instances and provides them with identity criteria), \mt{:Adult} and \mt{:Student} to be anti-rigid sortals (types that apply contingently to their instances and carry a principle of identify provided by a kind). Since \mt{:Adult} is a phase, it applies to its instances in virtue of some intrinsic aspects. Since \mt{:Student} is a role, it applies to its instances in virtue of some extrinsic (relational) aspects.

\begin{lstlisting}[language=SPARQL,label=lst:endurantypes,caption={An ontology fragment that defines metaproperties for Person, Adult and Student, and makes explicit the relational condition for Student.}]
:Person a gufo:Kind .

:Adult a gufo:Phase ;
       rdfs:subClassOf :Person .

:Student a gufo:Role ;
         rdfs:subClassOf :Person .

# relational condition for the :Student role

:Enrollment a owl:Class ;
            rdfs:subClassOf gufo:Relator .

:enrollmentInvolvesStudent a owl:ObjectProperty ;
                           rdfs:subPropertyOf gufo:mediates ;
                           rdfs:domain :Enrollment ;
                           rdfs:range :Student .

:Student rdfs:subClassOf [
            a owl:Restriction ;
            owl:onProperty [ owl:inverseOf :enrollmentInvolvesStudent ] ;
            owl:someValuesFrom :Enrollment
            ] .

\end{lstlisting}
    
These declarations allow tools to detect representational mistakes, e.g., that it is invalid: for a \mt{gufo:Kind} to be a sub-class of a \mt{gufo:Phase} or \mt{gufo:Role}; for an object to be an instance of more than one \mt{gufo:Kind}; for a \mt{gufo:Kind} to specialize any other sortal that specializes another \mt{gufo:Kind}, etc. See Section \ref{sec:constraints} for rules in SHACL that can detect these mistakes in a gUFO-based taxonomy and \cite{endurant_types_in_ontology_driven_conceptual_modeling__towards_ontouml_2_0_2018,Guizzardi2005} for formalization. 

The representation strategy employs OWL 2 punning, when a class is also treated as an instance of another class (in this case, \ttl{:Person a gufo:Kind} and, as defined in gUFO, \ttl{gufo:Kind a owl:Class}).

\subsection{High-order Types}
\label{sec:highorder}

A gUFO-based ontology may also specialize gUFO classes in the taxonomy of types, for example, in order to represent roles that persons play: \ttl{:PersonRole rdfs:subClassOf gufo:Role}.

In these cases, a user may want to establish explicitly the relation between \mt{:PersonRole} (the second-order type) and \mt{:Person} (the first-order type). For this purpose, the \mt{gufo:categorizes} property is provided. It identifies a \mt{gufo:Type} whose instances may be classified by instances of the categorizing higher-order type (Listing \ref{lst:highorder}).

The categorized type is termed the `base type' in the `powertype pattern' see \cite{multi_level_ontology_based_conceptual_modeling_2017}, the higher-order type is often called the `powertype'. The domain of \ttl{gufo:categorizes} is such that it excludes first-order types (instances of \ttl{gufo:AbstractIndividualType} or \ttl{gufo:ConcreteIndividualType}). In order words, a high-order type must not be a first-order type. 

\begin{lstlisting}[language=SPARQL,label=lst:highorder,caption=Person roles.]
:PersonRole gufo:categorizes :Person ;
            rdfs:subClassOf gufo:Role .

:Student   a :PersonRole ;
           rdfs:subClassOf :Person .
           
:Professor a :PersonRole ;
           rdfs:subClassOf :Person .
\end{lstlisting}

OWL 2 punning is used to capture the two facets of \mt{:Student} and \mt{:Professor} in this example: (i) as instances of \mt{:PersonRole}, and (ii) as subclasses of \mt{:Person}.

In another example (Listing \ref{lst:highorder2}), \ttl{:ShipType gufo:categorizes :Ship} and is a subclass of \mt{gufo:SubKind}. Instances of \mt{:ShipType} such as \mt{:Supercarrier} and \mt{:CargoShip} are declared subclasses of \mt{:Ship}.

\begin{lstlisting}[language=SPARQL,label=lst:highorder2,caption=Ship types.]
:ShipType gufo:categorizes :Ship ;
          rdfs:subClassOf gufo:SubKind .

:Supercarrier a :ShipType ;
              rdfs:subClassOf :Ship .
              
:CargoShip    a :ShipType ;
              rdfs:subClassOf :Ship .
\end{lstlisting}
        
\mt{gufo:categorizes} is the general (unspecific) form of categorization. The sub-property \mt{gufo:\-partitions} provides a more specific form, in which instances of the categorized type are classified by exactly one instance of the higher-order type.

For example, \ttl{:AnimalSpecies gufo:partitions :Animal}. Instances of \mt{:AnimalSpecies} such as \mt{:Lion}, \mt{:Hiena} must be disjoint subclasses of \mt{:Animal} (whether this is indeed the case is the subject of a SHACL constraint, see Section \ref{sec:mltconstraints}). Again, OWL 2 punning is used to capture the two facets of \mt{:Lion} and \mt{:Hiena} in this example: (i) as instances of \mt{:AnimalSpecies}, and (ii) as subclasses of \mt{:Animal} (Listing \ref{lst:partitioning}).

\begin{lstlisting}[language=SPARQL,label=lst:partitioning,caption=Animal species.]
:AnimalSpecies gufo:partitions :Animal .

:Hiena a :AnimalSpecies ;
       rdfs:subClassOf :Animal .
       
:Lion  a :AnimalSpecies ;
       rdfs:subClassOf :Animal ;
       owl:disjointWith :Hiena .

:Cecil a :Lion .
\end{lstlisting}

Note that the partitioned type (in the example \mt{:Animal}) may or may not be declared to be a disjoint union of the explicitly enumerated subclasses (such as \mt{:Lion}, \mt{:Hiena}). This is because other instances of the higher-order type (\mt{:AnimalSpecies}) may exist that are not explicitly enumerated in the ontology.

For further details and formalization of ``partitioning'', see \cite{multi_level_ontology_based_conceptual_modeling_2017} which combines UFO with MLT (a multi-level modeling theory).

\section{Original UFO Formalization and gUFO}
\label{sec:ufogufo}

As discussed in Guarino's seminal paper~\cite{guarino1998}, the conceptualization that an ontology aims to formally represent can be understood roughly as a set of ``intended world structures''. The quality of an ontology to rule out the \textit{unintended} world structures is called its \textit{precision} (while \textit{coverage} is its quality to allow, or ``rule in'', intended world structures)~\cite{guarino2004}. In principle, an ideal ontology would have models (in the logical sense) that correspond exactly to the intended world structures. Precision is particularly important in settings in which one wants to use the ontology as an artifact for meaning negotiation and consensus, i.e., as a \textit{reference ontology}. Because of this, ontologies (such as UFO, DOLCE, GFO, BFO) are typically formalized with languages that are more expressive than OWL-DL. In the case of UFO, first-order (alethic modal) logic was employed~\cite{ufo_unified_foundational_ontology_2021}.

OWL 2 DL (not unlike other formalisms) purposely establishes a number of constraints to expressiveness in order to guarantee certain computational properties (such as decidability). Losing expressiveness has the consequence that the same level of precision cannot be obtained. Some of us have explored this issue in great detail for the UFO-B fragment of UFO~\cite{representing_a_reference_foundational_ontology_of_events_in_sroiq_2019}. We have observed that maintaining the precision of a reference ontology when designing an implementation of it in \sroiq (which is the formal underpinning of OWL 2 DL) is a losing game. This is due to the loss of expressiveness given the constructs available in OWL and further due to incompatibilities that arise between the various \sroiq axioms as a result of the \sroiq \textit{structural restrictions} (concerning regularity and simplicity). \cite{representing_a_reference_foundational_ontology_of_events_in_sroiq_2019} shows that, considering all possible combinations of axioms that jointly satisfy the \sroiq restrictions while still maximizing precision, there are 12,288 UFO-B \sroiq theories that approximate the reference ontology. Each of these theories inevitably lose certain rules that were present in the reference formalization to ensure precision. Choosing one of these theories among thousands is far from trivial, and does not seem to have clear justifications in terms of ease of use, clarity, etc. The task involves seemingly arbitrary choices of which axioms to drop and which ones to keep, with little intuition offered to the users concerning the axioms that end up in the implementation. 

\begin{sloppypar}
We have aimed in gUFO for simplicity of implementation, with the following decisions concerning what formal aspects to retain:
\begin{enumerate}
\item The taxonomy axioms (as shown in Figure~\ref{fig:gufo-taxonomies}) have been captured directly, including completeness and disjointness declarations (using \ttl{owl:disjointWith}, \ttl{owl:disjointUnionOf} or \ttl{owl:AllDisjointClasses});
\item Domain, range and cardinality axioms for all object properties have been included, mirroring the reference ontology; 
\item Metaproperties of object properties (transitivity, asymmetry, irreflexivity, functionality constraints) have been declared when they do not violate the structural constraints of OWL.
\end{enumerate}
\end{sloppypar}

Examples of aspects that could not be retained include anti-transitivity of properties (which applies to \textit{inherence}), but cannot be expressed in OWL 2 DL; and some aspects of part-whole relations, as OWL 2 DL cannot formalize certain mereological notions as recognized in \cite{horrocks-i-2006-57-a}. In the case of the latter, they cannot be declared at the same time assymetric and transitive, as they would be considered \textit{non-simple} \cite{representing_a_reference_foundational_ontology_of_events_in_sroiq_2019} and, hence, they were declared only transitive (when applicable). (This is illustrative of the kind of `pick your poison' implementation decision with which OWL 2 DL implementers are faced. Declaring \textit{transitivity} allows reasoners to infer transitive proper parthood; declaring \textit{assymetry} would allow reasoners to flag the inconsistency when things are inadequately declared proper part of themselves.). Another example of a mereological constraint present in UFO but that cannot be represented in OWL 2 DL is weak supplementation \cite{bittner2007logical}.

An important aspect of UFO is its operationalization of OntoClean directives as part of its taxonomy of types. UFO can support the ontologist by ruling out ontologically incorrect taxonomies~\cite{types_and_taxonomic_structures_in_conceptual_modeling_2021} (for example, those in which a \textit{non-sortal} specializes a \textit{sortal} type; those in which a \textit{rigid} type specializes an \textit{anti-rigid} type; or, yet, those with classes that specialize more than one \textit{kind}.) However, those rules concern applications of \ttl{rdfs:subClassOf} and, hence, cannot be subject to axiomatization in OWL 2 DL. Given their centrality to UFO in the construction of high quality taxonomies, we have opted to complement the ontology implementation with SHACL constraints to enforce the missing UFO semantics. These constraints are discussed in detail in Section~\ref{sec:taxonomyconstraints} (see~\cite{types_and_taxonomic_structures_in_conceptual_modeling_2021} for similar SPARQL constraints). 

There are also UFO and MLT \cite{toward_a_well_founded_theory_for_multi_level_conceptual_modeling_2016, multi_level_ontology_based_conceptual_modeling_2017}  constraints which involve \ttl{rdfs:subClassOf} and \ttl{rdf:type}, and, thus, are also not subject to axiomatization in OWL 2 DL. Consider the patterns for the representation of high-order domain types discussed in Section~\ref{sec:highorder}. Declaring that \ttl{:ShipType gufo:categorizes :Ship} would imply that instances of \ttl{:ShipType} such as \ttl{:CargoShip} are subclasses of \ttl{:Ship}. Likewise, declaring that \ttl{:AnimalSpecies gufo:partitions :Animal} would imply that instances of \ttl{:AnimalSpecies} such as \ttl{:Lion} are subclasses of \ttl{:Animal}. In the latter case, the semantics of ``partitioning'' as specified in \cite{toward_a_well_founded_theory_for_multi_level_conceptual_modeling_2016, multi_level_ontology_based_conceptual_modeling_2017} would also imply that instances of \ttl{:AnimalSpecies} are mutually disjoint and that they together exhaust \ttl{:Animal}. All of those formal consequences of UFO and MLT for instances of high-order types are not directly implemented in gUFO. Again, they are made available as SHACL/SPARQL constraints, which are presented in Section~\ref{sec:mltconstraints}.

In addition to the matters of expressiveness, there are other differences of gUFO to the original formalization of the reference ontology that stem from: (i) the treatment of modality; (ii) scope; (iii) rules that are already enforced by the semantics of OWL. These are discussed in the sequel. 

With respect to the treatment of modality, the formalization of UFO accounts for the fact that an object may instantiate a class in a certain world and not in another world (while some other classes apply necessarily to their instances). Take, for instance, the classes Teenager and Professor. Someone is contingently a Teenager (or  Professor), and thus we can conceive of a possible world in which they exist but do not instantiate that class (e.g., when they were just born). These are called anti-rigid classes. The same cannot be said of the class Person. A person is necessarily a person in every possible world in which they exist. This is accounted for  in~\cite{ufo_unified_foundational_ontology_2021} by using the standard modal operators ($\Box$ and $\Diamond$) with a possibilist interpretation. In constrast, OWL has no native support for modalities, and \ttl{rdf:type} is not time- or world-indexed; classes are considered sets (thus with no changes in membership). Hence, we cannot distinguish the mode of instantiation that apply to \ttl{:Paul rdf:type :Person} from that which applies to \ttl{:Paul rdf:type :Teenager}. The solution in gUFO to preserve the support for non-rigid classes was to reify \textit{situations} as discussed in \ref{sec:situations}. The same applies to other aspects of change beyond classification, which are dealt with situations, as also discussed in that section (change in the value of qualities, change in mereological relations, etc.) In that way, gUFO offers a regular representation strategy for historical or current states of affairs, in an otherwise time-neutral formalism (OWL).

A couple of adjustments of scope are documented in the gUFO specification (in the form of \ttl{rdf:comment} annotations). UFO originally included the more abstract notion of ``Substantial'' \cite{Guizzardi2005}, which generalizes both objects and amounts of matter. That notion was left out from gUFO, together with the notion of amount of matter. The reason is that, as discussed in \cite{Guizzardi2005}, ``non-object substantials (amounts of matter) can only be represented in a conceptual model as quantities.'' Hence, for all practical purposes, there is no impact of the conceptual simplification, as gUFO supports the representation of maximally-self-connected amounts of matter through \ttl{gufo:Quantity} as a subclass of \ttl{gufo:Object}. This also contributes to minimizing the use of philosophical jargon in gUFO (\textit{substantial} in this case). The original UFO literature also employed the term `\textit{Moment}', which stems from the German `Momente' in the writings of Husserl~\cite{Guizzardi2005}. This was replaced by the more common term `Aspect' in gUFO, as the original term not only is jargon, but perversely evokes the sense found in the dictionary as `a very brief period of time', which is certainly not what was aimed for. The term `trope' is also avoided altogether. The term `Universal' used in some UFO literature was replaced with the more common `type'. `Quale' was replaced with the more descriptive \textit{quality value}. `Perdurant' is `Event' (which was a term already favored in UFO-B literature~\cite{representing_a_reference_foundational_ontology_of_events_in_sroiq_2019,towards_ontological_foundations_for_the_conceptual_modeling_of_events_2013}).

For simplicity, we have left out the notion of `Set' and its specializations `Quality Structure', `Quality Dimension' and `Quality Space'. In the case of simple quality structures, these are addressed directly through data sub-properties of \ttl{gufo:hasQualityValue} with specific datatype ranges. In the case of multidimensional quality spaces, gUFO supports the reification of quality values (\ttl{gufo:hasReifiedQualityValue}) with abstract individuals that have value components (through \ttl{gufo:hasValueComponent}). 

Since gUFO is intended to be used in knowledge graph implementations, a number of decisions concerning the representation of time in practical settings were established. UFO-B establishes that time points are totally ordered~\cite{representing_a_reference_foundational_ontology_of_events_in_sroiq_2019}, but is silent with respect to the granularity and representation of time instants and intervals in calendars. As presented in Section \ref{sec:taxonomy_of_invididuals}, data properties with well supported XML Schema datatypes were included to standardize a simple treatment of time (with begin and end points in two possible granularities). Also as indicated there, more sophisticated uses are made possible with the reuse of OWL-Time~\cite{owltime} \mt{time:Instant}. %

Finally, there are a few of axioms in the original formalization of UFO that are already natively encoded in the semantics of OWL. For example, the UFO axioms concerning the definition of subclassing are already encoded in the semantics of \ttl{rdfs:subClassOf}. %
Similarly, in OWL there is native support for the distinction between data values in a \textit{data domain} and the \textit{object domain}~\cite{owl2directsemantics}. So, unlike the original formalization of UFO, there is no need to consider data values as abstract individuals.

\section{Preserving Ontological Constraints beyond OWL 2 DL}
\label{sec:constraints} 

As discussed in the previous sections, a number of important consequences of the UFO and MLT original formalizations cannot be directly encoded in OWL 2 DL. In this section, we show how they can be implemented using SHACL and SPARQL. 

\subsection{Shapes for the Taxonomy of Types}
\label{sec:taxonomyconstraints}

The metaproperties of the typology of types allows us to provide a number of `shapes' for a gUFO-based knowledge graph. These shapes are specified in SHACL. They reflect semantic constraints which follow from the UFO axiomatization~\cite{ufo_unified_foundational_ontology_2021}. %
These constraints can be checked automatically, operationalizing the guidelines in the OntoClean methodology for taxonomies~\cite{Guarino2004ontoclean}. 

For example, the shape in Listing \ref{lst:antirigid} ensures that rigid and semi-rigid types cannot specialize anti-rigid types~\cite{ufo_unified_foundational_ontology_2021}. This is a theorem in UFO, and follows logically from the definition of rigidity, anti-rigidity, and semi-rigidity in the original axiomatization (see proof in~\cite{endurant_types_in_ontology_driven_conceptual_modeling__towards_ontouml_2_0_2018}). 
We have made reference in the shape specification to both the `leaf' classes in the taxonomy of types (\mt{gufo:Category}, \mt{gufo:Category}, \mt{gufo:Role}, etc.) and their subsuming ones (\mt{gufo:RigidType}, \mt{gufo:SemiRigidType}, \mt{gufo:AntiRigidType}, in order to make sure that various configurations of gUFO-based ontologies are accounted for.

\begin{lstlisting}[language=SPARQL,label=lst:antirigid,caption=SHACL constraint to ensure that rigid and semi-rigid types do not specialize anti-rigid ones.,breaklines=true,showstringspaces=false]
<https://w3id.org/nemo/gufoshapes#RigidOrSemiRigidShape> a sh:NodeShape ;
  sh:targetClass gufo:RigidType, gufo:Category, gufo:Kind, gufo:SubKind, gufo:SemiRigidType, gufo:Mixin ;
  sh:property [
    sh:name "Forbidden anti-rigid type specialization" ;
    sh:message "Rigid and semi-rigid types can't specialize anti-rigid types." ;
    sh:path rdfs:subClassOf ;
    sh:not [
      sh:or (
        [ sh:class gufo:AntiRigidType ]
        [ sh:class gufo:Phase ]
        [ sh:class gufo:PhaseMixin ]
        [ sh:class gufo:Role ]
        [ sh:class gufo:RoleMixin ]
      )
    ]
  ]
  .
\end{lstlisting}

Listing \ref{lst:sortality} captures the rule that non-sortal types cannot specialize sortal types. This is because sortals either provide a principle of identity to their instances, or inherit one from another sortal that does. Instances of a non-sortal type are not restricted to a common principle of identity, which would be enforced (incorrectly) by a specialization violating this rule.
 
\begin{lstlisting}[language=SPARQL,label=lst:sortality,caption=SHACL constraint to ensure that non-sortal types do not specialize sortal types.,breaklines=true,showstringspaces=false]
<https://w3id.org/nemo/gufoshapes#NonSortalShape> a sh:NodeShape ;
  sh:targetClass gufo:NonSortal, gufo:Category, gufo:PhaseMixin, gufo:RoleMixin, gufo:Mixin ;
  sh:property [
    sh:name "Forbidden sortal specialization" ;
    sh:message "Non-Sortal types can't specialize Sortal types." ;
    sh:path rdfs:subClassOf ;
    sh:not [
      sh:or (
        [ sh:class gufo:Sortal ]
        [ sh:class gufo:Kind ]
        [ sh:class gufo:SubKind ]
        [ sh:class gufo:Phase ]
        [ sh:class gufo:Role ]
      )
    ]
  ] ;
  .
\end{lstlisting}

Listing \ref{lst:kindexists} ensures that every sortal that is not a kind carries a principle of identify that it inherits from another sortal it specializes. Note that this kind of constraint is applied to a complete ontology, and that it may not be enforced when using an open-world assumption. 

\begin{lstlisting}[language=SPARQL,label=lst:kindexists,caption=SHACL constraint to ensure that an identity-carrying sortal inherits identity criteria from another sortal.,breaklines=true,showstringspaces=false]
<https://w3id.org/nemo/gufoshapes#BaseSortalShape> a sh:NodeShape ;
  sh:targetClass gufo:SubKind, gufo:Phase, gufo:Role ;
  sh:property [
    sh:name "Missing identity provider" ;
    sh:message "Sortal types must specialize a kind or some other sortal." ;
    sh:path rdfs:subClassOf ;
    sh:qualifiedValueShape [
      sh:or (
        [ sh:class gufo:Kind ]
        [ sh:class gufo:SubKind ]
        [ sh:class gufo:Phase ]
        [ sh:class gufo:Role ]
      )
    ] ;
    sh:qualifiedMinCount 1 ;
  ]
  .
\end{lstlisting}

Listing \ref{lst:kindcantspecialize} ensures that no kind specializes another (identity-carrying) sortal, i.e., that it is the `ultimate' sortal in a taxonomy.

\begin{lstlisting}[language=SPARQL,label=lst:kindcantspecialize,caption=SHACL constraint to ensure that a kind is the `ultimate' sortal.,breaklines=true,showstringspaces=false]
<https://w3id.org/nemo/gufoshapes#KindShape> a sh:NodeShape ;
  sh:targetClass gufo:Kind ;
  sh:property [
    sh:name "Forbidden sortal specialization" ;
    sh:message "Kinds cannot specialize sortal types (i.e., types that already set or inherit an identity principle)." ;
    sh:path rdfs:subClassOf ;
    sh:not [
      sh:or (
        [ sh:class gufo:SubKind ]
        [ sh:class gufo:Phase ]
        [ sh:class gufo:Role ]
      )
    ]
  ]
  .
\end{lstlisting}

Listing \ref{lst:powertype} ensures that instances of \mt{gufo:EndurantType} and its subclasses do not specialize classes that are disjoint from \mt{gufo:Endurant}. \mt{gufo:EndurantType} is defined as a high-order type whose instances are types of endurants. Hence, they cannot be specializations of types of events, of abstract individuals, or of situations. Such specializations would otherwise necessarily have an empty extension.
\begin{lstlisting}[language=SPARQL,label=lst:powertype,caption=SHACL constraint to ensure that the endurant types do not specialize disjoint non-endurant types.,breaklines=true,showstringspaces=false]
<https://w3id.org/nemo/gufoshapes#EndurantTypeShape> a sh:NodeShape ;
  sh:targetClass gufo:EndurantType, gufo:RigidType, gufo:NonRigidType, 
    gufo:AntiRigidType, gufo:SemiRigidType, gufo:Phase, gufo:PhaseMixin,
    gufo:Role, gufo:RoleMixin, gufo:Mixin, gufo:NonSortal, gufo:Category,
    gufo:Sortal, gufo:Kind, gufo:SubKind ;
  sh:property [
    sh:name "Powertype pattern violation" ;
    sh:message "Endurant types cannot specialize classes disjoint from Endurant." ;
    sh:path rdfs:subClassOf ;
    sh:not [
      sh:in (
        gufo:AbstractIndividual
        gufo:QualityValue
        time:Instant
        gufo:Event
        gufo:Participation
        gufo:Situation
        gufo:QualityValueAttributionSituation
        gufo:TemporaryConstitutionSituation
        gufo:TemporaryInstantiationSituation
        gufo:TemporaryParthoodSituation
        gufo:TemporaryRelationshipSituation
      )
    ]
  ]
  .
\end{lstlisting}

Rules corresponding to these constraints have been implemented into a Protégé plugin.\footnote{\url{https://github.com/nemo-ufes/ufo-protege-plugin}} It adds a `File $\rightarrow$ Validate GUFO rules' menu item to Protégé, with results displayed in a specialized tab  (available under `Window $\rightarrow$ Tabs $\rightarrow$ UFO Validation Tab').

\subsection{Shapes for High-Order Types and their Base Types}
\label{sec:mltconstraints}

Listing~\ref{lst:categorization} shows the constraint that enforces the semantics of \textit{categorization} between a high-order type a base type. It selects those instances of a categorizing high-order type that are not declared as subclasses of the base type. (In the listing, we have omitted the prefixes for brevity, see line~7). 

\begin{lstlisting}[language=SPARQL,label=lst:categorization,caption=SPARQL-based SHACL constraint to enforce the semantics of categorization.,breaklines=true,showstringspaces=false,numbers=left]
<https://w3id.org/nemo/gufoshapes#CategorizationShape> a sh:NodeShape ;
    sh:targetSubjectsOf gufo:categorizes ;
    sh:sparql [
        a sh:SPARQLConstraint ;
        sh:message "Instances of a categorizing higher-order type (focus node) must be subclasses of the categorized base type." ;
        sh:prefixes [
            # ommitted for brevity
            ]
        ] ;
        sh:select """
            SELECT $this ?instance
            WHERE {
                $this gufo:categorizes ?baseType .
                ?instance rdf:type $this .
                FILTER NOT EXISTS { ?instance rdfs:subClassOf ?baseType }
            }
        """ ;
    ] .

\end{lstlisting}

Listing \ref{lst:partitioning2} checks whether the instances of a \textit{partitioning} type are disjoint. It checks for each pair of instances whether they are: (i) declared pairwise disjoint with \ttl{owl:disjointWith}; (ii) members of the same \ttl{owl:AllDisjointClasses} list; or, (iii) members of the same \ttl{owl:disjointUnionOf} list. Since \textit{partitioning} is a special form of \textit{categorization} (and \ttl{gufo:partitions} is a subproperty of \ttl{gufo:categorizes}), the constraint in the previous listing also applies to instances of a partitioning high-order type, and hence, they are required to be subclasses of the partitioned base type.
 
\begin{lstlisting}[language=SPARQL,label=lst:partitioning2,caption=SPARQL-based SHACL constraint to enforce the semantics of partitioning.,breaklines=true,showstringspaces=false,numbers=left]
<https://w3id.org/nemo/gufoshapes#PartitionPairwiseDisjointnessShape> a sh:NodeShape ;
    sh:targetSubjectsOf gufo:partitions ;
    sh:sparql [
        a sh:SPARQLConstraint ;
        sh:message "Instances {?type1} and {?type2} of the partitioning type {$this} are not declared disjoint." ;
        sh:prefixes [
            # ommited for brevity
        ] ;
        sh:select """
            SELECT $this ?type1 ?type2
            WHERE {
                $this gufo:partitions ?baseType .                
                # Select two instances of the HoT
                ?type1 rdf:type $this .
                ?type2 rdf:type $this .
                FILTER (?type1 != ?type2)
                # Filter out classes that are semantically equivalent
                FILTER NOT EXISTS { ?type1 owl:equivalentClass ?type2 }
                FILTER NOT EXISTS { ?type1 owl:sameAs ?type2 }

                # Filter out pairs that ARE declared disjoint
                FILTER NOT EXISTS {
                    { ?type1 owl:disjointWith ?type2 }
                    UNION
                    # members of the same owl:AllDisjointClasses list
                    {
                        ?adc rdf:type owl:AllDisjointClasses ;
                             owl:members ?members .
                        ?members rdf:rest*/rdf:first ?type1 .
                        ?members rdf:rest*/rdf:first ?type2 .
                    }
                    UNION
                    # members of the same owl:disjointUnionOf list
                    {
                        ?cls owl:disjointUnionOf ?unionList .
                        ?unionList rdf:rest*/rdf:first ?type1 .
                        ?unionList rdf:rest*/rdf:first ?type2 .
                    }
                }
            }
        """ ;
    ] .
\end{lstlisting}

\section{Assessment}
\label{sec:assessment}

\subsection{Consistency}

gUFO was found consistent using HermiT~\cite{DBLP:conf/owled/ShearerMH08} version 1.4.3.456 integrated in Protégé. We have checked for pitfalls and anti-patterns with the online {Oops! tool}~\cite{poveda2014oops} (using the RDF/XML serialization for gUFO, which is required by the tool). The only supposed `pitfalls' detected were: `P04: Creating unconnected ontology elements. 1 case', `Results for P11: Missing domain or range in properties. 4 cases.' and `P13 - Inverse relationships not explicitly declared. 34 cases'. All of those, however, are the result of intentional design choices: The only  supposedly `unconnected element' is \mt{gufo:Individual}, which is a key element in the taxonomy backbone of gUFO, and is declared disjoint with \mt{gufo:Type}, creating an important distinction in the ontology. The `missing' domain or range declarations refer to: in one case, a property whose superproperty has a range declaration, and hence, this declaration is inherited (clearly the pitfall detection tool does not perform the required inference in this case.) In the other 3 cases, the missing range declarations concern the general data properties (such as \mt{gufo:hasQualityValue}). Given the level of generality of the foundational ontology, this intentional omission is justified. Finally, we have opted not to include inverse object properties for the sake of simplicity. %

\subsection{FAIRness}

Rich metadata and provenance information has been provided as part of gUFO with annotations. As part of the ontology evaluation procedure, we have run the FOOPS! Ontology Pitfall Scanner for the FAIR principles~\cite{foops2021}, which results in very high FAIRness overall score (92\%) (the only thing barring a 100\% score is the choice of detailed provenance metadata that the tool considers; we have provided the same metadata using different Dublin Core properties).

Label and comment annotations (\mt{rdfs:label} and \mt{rdfs:comment}) are available for all gUFO elements. The documentation includes a cross-linked reference\footnote{\url{https://purl.archive.org/nemo/doc/gufo\#toc}} with definitions following the Aristotelian \textit{genus} and \textit{differentia} style for all elements. It is generated automatically from the ontology annotations using a (lightly) customized version of pyLODE.\footnote{\url{https://github.com/RDFLib/pyLODE}}

\section{Community Adoption}
\label{sec:community-adoption}

Since its proposal, gUFO has supported the operationalization of ontologies in several academic and industry settings. In this section, we present a diverse set of third-party works exemplifying adoption in a variety of domain and use case scenarios.

For example, in the Enterprise Architecture domain, Pinheiro et al. \cite{pinheiro2024lightweight} have employed gUFO in the construction of a knowledge graph designed to hold information mined from information systems API communication enabling mining of enterprise architecture models representing the architecture connecting those systems. In his turn, Schieferdecker \cite{schieferdecker2025software} has employed gUFO to develop a well-defined taxonomy of AI-augmentation techniques in software testing (\textit{ai4st}). Nascimento and Oliveira \cite{nascimento2023ontology}
proposed SpaceCOn, a gUFO-based ontology for smart environments and context-aware applications, which is applicable in the publication of heterogeneous information on the Semantic Web as well as in the support of tailored information systems. In \cite{melo2025analise}, Melo used a gUFO-based ontology to construct and analyze a dataset on school evasion in higher-education and produce insights in the impact of governmental policies to support students at risk of evasion. Antunes and colleagues \cite{cauadoi:10.1177/15705838251383667} have used gUFO in the development of an implementation of their case on oil flow inside a reservoir, enriching it a with additional domain-specific SWRL rules. Ubbiali and colleagues have developed the Sustainability Core Ontology (SCO) \cite{ubbiali2025sustainability} offering an alignment with both gUFO (SCO-U) and BFO (SCO-B), which represents an interesting comparison point to both gUFO and BFO communities. 

A number of OntoUML ontologies have also benefited from automated transformation into gUFO-based OWL implementations \cite{ontology_driven_conceptual_modeling_as_a_service_2021}. This is the case of the SEBIM ontology \cite{silva2023conceptual} on Civil Engineering applications and sustainability certification processes. This operationalization strategy gives access to Semantic Web technologies to users of the proposed domain ontology. This strategy is also employed by Demori \cite{demori2023abordagem} in his OntoUML ontology on military operations, as well as Kom\'{a}romi \cite{komaromi2023philosophy} in his series of OntoUML ontologies on philosophical theories. 

In industry and government, other publicly available sources using gUFO include: 
\begin{itemize}
    \item the Crime Against Children (CAC) Ontology (published by Project VIC International), which provides a gUFO-based comprehensive framework for modeling child exploitation investigations and digital forensics;\footnote{\url{https://site.cacontology.projectvic.org/}}
    \item the ONZ-G Ontology \cite{segers2026ontologie} (published by the Dutch Zorginsituut Nederland) to provide semantical alignment for important domain concepts with gUFO definitions connected via SKOS properties \cite{miles2009skos};
    \item the gUFO profile of the Unified Cyber Ontology (UCO),\footnote{\url{https://github.com/ucoProject/UCO-Profile-gufo}} as part of a Linux Foundation Project (and an accompanying extensive set of SHACL constraints for gUFO\footnote{\url{https://github.com/Cyber-Domain-Ontology/CDO-Shapes-gufo}});%
    \item the Cameo Concept Modeler tool from No Magic, which includes a diagrammed reference concept model for gUFO in its CCM library since 2022. The tool also includes a template to create gUFO-based concept modeling projects.\footnote{\url{https://docs.nomagic.com/spaces/CCMP2021xR2/pages/82753847/2021x+Refresh2+Version+News}}
\end{itemize}

More recently, UFO has been approved for publication as a Draft International Standard in the International Organization for Standardization (as ISO/IEC 21838-5 \cite{iso218385}). As part of the requirements for the standardization of top-level ontologies, ISO/IEC 21838-1 \cite{iso218381} established that ontology standards must include a formalization in OWL. The gUFO formalization was included as an integral part of the standard. We expect that the progress of the ISO/IEC standardization will further boost the adoption of gUFO.

\section{Related Work}
\label{sec:related-work}

Here, we compare gUFO with the simplified OWL versions of foundational ontologies, in particular, DOLCE+DnS Ultralite (DUL), and BFO.

UFO traces part of its history to DOLCE. Hence, there is quite some alignment in a number of top-level distinctions in gUFO and DUL, with both (3D) ontologies covering objects, events, situations and types. Both allow the reification of qualities. Nevertheless, unlike gUFO, DUL does not reify relational aspects (nor their parts). DUL, despite evoking `ultra lite' in its name, is significantly larger than gUFO, with 79 classes as opposed to 51, and 113 object properties as opposed to 40. This can be traced to its inclusion of a number of classes in the `social' domain such as norm, contract, agent, project, theory, method, plan, workflow, workflow execution, narrative, etc. In UFO, these have been factored out into UFO-C~\cite{DBLP:conf/cibse/GuizzardiFG08}, which is not part of gUFO. Differently from gUFO, DUL does not cover the metaproperties of types, although it reifies types with its notion of \mt{Concept}. It does not support the powertype variants that gUFO supports. An analysis of DUL with Oops! reveals a number of missing annotations (22 cases); license metadata is not provided. (FOOPS! could not load DUL; it is unclear why.)

BFO, similarly to UFO and DOLCE is a 3D-ontology and also distinguishes objects and events, and accounts for the participation of objects into events. In contrast with both DUL and gUFO, BFO employs (philosophical) jargon extensively for those distinctions (e.g., `occurrent', `continuant') and for many others (`generically-dependent continuant', `specifically-dependent continuant', `continuant fiat boundary', etc.) We argue this has significant consequences for BFO's ease of use. In contrast with DUL and gUFO, BFO (and its OWL implementation) does not support types at all (and then of course, it does not support patterns for high-order types, nor the detection of problematic taxonomies). It also does not admit abstract entities (such as sets, numbers). This seems to stem from BFO's explicit ``realist'' metaphysical stance, and could pose a challenge for applications in certain domains (the social domain included). 

The object properties in BFO's implementation (such as \mt{continuant part of}) are defined in such a way that they relate two entities in case there is a time in which the relation holds (``$b$ \mt{continuant part of} $c$ $=_{Def}$ $b$ and $c$ are continuants \& there is some time $t$ such that $b$ and $c$ exist at $t$ \& $b$ continuant part of $c$ at $t$''.) A separate ontology with object properties that take into account time/change is offered, but these are not provided in the form of situations with temporal boundaries (like in gUFO and DUL), but instead with temporalized variants for the object properties, with two semantics: applying at ``all times'' and applying at ``some time'' (for example, distinguishing \mt{has continuant part at all times} and \mt{has continuant part at some time}.) The specific times during which the temporary parthood applies cannot be expressed (unlike in DUL and gUFO).  An analysis of BFO's implementation with Oops! reveals a number of missing annotations (35 cases), and some other minor warnings.

\section{Final Considerations}
\label{sec:conclusions}
In this paper, we have presented the gUFO implementation of the Unified Foundational Ontology.
gUFO is a foundational ontology implementation that allows for a number of scenarios for reuse in the construction of RDF/OWL knowledge graphs and ontologies with different levels of sophistication. It emphasizes freedom from jargon, adherence to Semantic Web community practices. It distinguishes itself from other foundational ontology implementations for its quality characteristics, but also for its support for a typology of types, support for high-order types, and reification patterns which solve recurrent problems that domain ontology developers face when adopting Semantic Web technologies. The resource is well-documented, available in a stable location, and includes rich metadata in line with the FAIR principles. gUFO and its supporting tools are maintained by a collaboration between the Federal University of Espírito Santo in Brazil and the University of Twente, in The Netherlands, currently with funding from Brazilian state agencies (FAPES and CNPq). Github issue handling and discussion functionality in the project repository\footnote{\url{https://github.com/nemo-ufes/gufo}} are used to provide community feedback and support.

In the future, we will develop a number of gUFO-based core ontology implementations. We expect to develop implementations for detailed event/process representations, for specialized agentive, social and intentional notions~\cite{DBLP:conf/cibse/GuizzardiFG08}, and for legal aspects~\cite{towards_a_legal_core_ontology_based_on_alexy_s_theory_of_fundamental_rights_2015}. 

gUFO is well-integrated into the tool ecosystem for conceptual modeling based on UFO, including the OntoUML language and its tools~\cite{ontology_driven_conceptual_modeling_as_a_service_2021,towards_ontological_foundations_for_conceptual_modeling__the_unified_foundational_ontology__ufo__story_2015}. Presently, the OntoUML Visual Paradigm plugin\footnote{\url{https://github.com/OntoUML/ontouml-vp-plugin}} includes a transformation from OntoUML into gUFO-based OWL ontologies. The editor for the corresponding textual notation for OntoUML\,--\,named Tonto\footnote{\url{https://github.com/matheuslenke/Tonto}}\,--\,also includes a transformation of specifications in the language to gUFO-based OWL ontologies. This transformation has been used to obtain a catalog of gUFO-based ontologies\footnote{\url{https://github.com/matheuslenke/tonto-catalog}} derived from the OntoUML catalog~\cite{a_fair_catalog_of_ontology_driven_conceptual_models_2023}. As gUFO adoption grows, the catalog will include further gUFO `native' ontologies (such as some of those third-party developments mentioned in Section \ref{sec:community-adoption}).

\section*{Acknowledgements}
The authors would like to recognize the contributions of Ricardo de Almeida Falbo (\textit{in memoriam}) to the development of gUFO. They would also like to thank Matheus Lenke Coutinho for his work on the repository of gUFO-based ontologies. We would also like to thank anonymous reviewers of a previous version of this paper and Jim Logan for suggesting corrections. This work has been supported by CNPq (443130/2023-0, 313412/2023-5) and FAPES (1022/2022).

\bibliography{references}

\end{document}